\documentclass{article}

\usepackage[preprint]{neurips_2024}

\usepackage[utf8]{inputenc} 
\usepackage[T1]{fontenc}    
\usepackage{hyperref}       
\usepackage{url}            
\usepackage{booktabs}       
\usepackage{amsfonts}       
\usepackage{nicefrac}       
\usepackage{microtype}      
\usepackage{graphicx}
\usepackage{subcaption} 
\usepackage{multirow}
\usepackage[dvipsnames]{xcolor}
\usepackage{amsmath}

\title{Synthetic imagery for fuzzy object detection: A comparative study}

\author{
  Siavash H. Khajavi \\
  Department of Industrial Engineering and Management\\
  Aalto University\\
  \texttt{siavash.khajavi@aalto.fi}\\
  \And 
  Mehdi Moshtaghi \\
  Department of Computer Science \\
  Aalto University \\
  \texttt{mehdi.moshtaghi@aalto.fi} \\
  \And 
  Dikai Yu \\
  Department of Design \\
  Aalto University \\
  \texttt{dikai.yu@aalto.fi} \\
  \And 
  Zixuan Liu \\
  Department of Computer Science \\
  Tulane University \\
  \texttt{zliu41@tulane.edu} \\
  \And 
   Kary Främling \\
   Department of Industrial Engineering and Management \\
   Aalto University \\
   \texttt{kary.framling@aalto.fi}
   \And 
   Jan Holmström \\
   Department of Industrial Engineering and Management\\
  Aalto University\\
  \texttt{jan.holmstrom@aalto.fi}
}

\begin{document}

\maketitle

\begin{abstract}
  The fuzzy object detection is a challenging field of research in computer vision (CV). Distinguishing between fuzzy and non-fuzzy object detection in CV is important. Fuzzy objects such as fire, smoke, mist, and steam present significantly greater complexities in terms of visual features, blurred edges, varying shapes, opacity, and volume compared to non-fuzzy objects such as trees and cars. Collection of a balanced and diverse dataset and accurate annotation is crucial to achieve better ML models for fuzzy objects, however, the task of collection and annotation is still highly manual. In this research, we propose and leverage an alternative method of generating and automatically annotating fully synthetic fire images based on 3D models for training an object detection model. Moreover, the performance, and efficiency of the trained ML models on synthetic images is compared with ML models trained on real imagery and mixed imagery. Findings proved the effectiveness of the synthetic data for fire detection, while the performance improves as the test dataset covers a broader spectrum of real fires. Our findings illustrates that when synthetic imagery and real imagery is utilized in a mixed training set the resulting ML model outperforms models trained on real imagery as well as models trained on synthetic imagery for detection of a broad spectrum of fires. The proposed method for automating the annotation of synthetic fuzzy objects imagery carries substantial implications for reducing both time and cost in creating computer vision models specifically tailored for detecting fuzzy objects.
\end{abstract}

\section{Introduction}
\label{intro}
Fuzzy objects, such as fire, smoke, are common in natural physical life.  Detecting and locating such objects in the images remains difficult in the computer vision field due to their blurred edges, variable shapes and varying intensity levels compared to well-defined objects. 

The success of AlexNet~\cite{krizhevsky2012imagenet} on ImageNet~\cite{deng2009imagenet} led to the use of deep convolutional networks (CNNs) for fuzzy object detection. Zhang et al.~\cite{zhang2016deep} proposed a fine-tuned pre-trained AlexNet model for forest fire detection. Muhammad et al. fine-tuned different variants of CNNs like AlexNet~\cite{muhammad2018early}, SqueezeNet~\cite{muhammad2018efficient},GoogleNet~\cite{muhammad2018convolutional}, and MobileNetV2~\cite{muhammad2019efficient} for detection. However, the performance of their methods is not promising because the datasets they used is small and limited. They used Foggia’s dataset~\cite{foggia2015real} which only contains 14 fire and 17 non-fire videos, and the frames of the videos are highly similar, which greatly restricts the performance of the model. It is known that CNNs methods are only effective when training on large scale datasets. For example, the well-used ImageNet ILSRVC dataset~\cite{deng2009imagenet} contains 1 million images for training. Obtaining such vast amounts of training datasets with real-world data of fuzzy objects, especially for fire and smoke, is challenging because of their sporadic nature and the difficulty in capturing high-quality data under hazardous conditions. Moreover, publicly-available and high-quality datasets of fire and smoke are limited; for example, the D-Fire dataset~\cite{de2022automatic}, the largest published fire dataset, comprises 21,000 annotated fire and smoke images, which is not comparable with ImageNet. Another issue with D-Fire is that a significant portion of the dataset contains sequential frames of surveillance videos, which are highly similar images. After removing duplicate and similar images, based on their VGG \cite{simonyan2015vgg} embeddings, approximately half of the original dataset remains. This also makes D-Fire an inappropriate dataset as a benchmark to test on. As a result, current methods perform poorly on fuzzy object detection. Generating large, realistic, and diverse data of fire and smoke is vital for developing effective detection algorithms to detect fires.

In order to mitigate the requirement of large datasets, synthetic data has been proposed to solve this challenge and be used for training CNN models. Researchers at OpenAI~\cite{tobin2017domain} use domain randomization, which is a technique for training models on simulated images that transfer to real images by randomizing rendering in the simulator. Their experiments show that, with enough variability in the simulator, the deep neural network trained only on simulated images can be successfully transferred to real-world for object localization. In addition, researchers at Google~\cite{hinterstoisser2019annotation} have successfully demonstrated the efficacy of synthetic data for non-fuzzy object detection. They leverage a large dataset of 3D background models and densely render them using full domain randomization. This yields background images with realistic shapes and texture, on top of which they render the objects of interest. Moreover, Unity~\cite{borkman2021unity} published free tools called “unity computer vision” and “unity simulation” for generating large-scale datasets with labels, randomizers, samples, and rendering capabilities. However, all their works focus on rendering limited non-fuzzy objects with well-defined boundaries or characteristics, such as boxes, bottles, books. In this research, we intend to expand this to fuzzy objects and propose a method of 3D model-based synthetic imagery generation for this purpose. Our research questions are as follows:

\begin{enumerate}
    \item Is the fully synthetic data generated by 3D model-based synthetic imagery generation method effective in enabling a computer vision ML model to detect fire? 
    \item How to implement automated annotation for fuzzy objects in the 3D model-based synthetic imagery data?
    \item What is the comparative performance of the ML model trained on fully synthetic imagery, the ML model trained on real imagery, and the ML model trained on mixed imagery in the detection of fire?
\end{enumerate}

 

\section{Related Work}
The computer vision field has allocated substantial resources towards the development of datasets like PASCAL VOC~\cite{everingham2010pascal}, MS COCO~\cite{lin2015microsoft}, and ImageNet~\cite{deng2009imagenet}, which have significantly promoted the research forward in complex areas such as image classification~\cite{he2016deep}, object detection~\cite{ren2015faster,redmon2016you,lin2017focal}, and semantic segmentation~\cite{he2017mask, long2015fully}. However, these datasets do not encompass all scenarios of interest to researchers, nor do they facilitate the creation of new datasets, especially some domain-specific datasets. Notice that acquiring and manually annotating such domain-specific datasets can be costly, labor-intensive, and prone to mistakes~\cite{hinterstoisser2019annotation}. A widely adopted strategy to enhance the efficacy of detection algorithms involves augmenting real datasets with synthetic data~\cite{su2015render,georgakis2017synthesizing}. To bridge the gap between synthetic and real image characteristics, various techniques have been investigated~\cite{shrivastava2017learning,hinterstoisser2019annotation, inoue2018transfer}. Shrivastava et al.~\cite{shrivastava2017learning} propose a Simulated+Unsupervised (S+U) learning method, employing adopted Generative Adversarial Networks (GANs) to align the distribution of synthetic and real images more closely. Hinterstoisser et al.~\cite{hinterstoisser2019annotation} deploy a technique where background scenes are rendered with complete domain randomization~\cite{tobin2017domain}, incorporating realistic shapes and textures. On top of these backgrounds, the target objects of interest are then rendered, encompassing a comprehensive array of poses and conditions to ensure a thorough representation in training datasets. Furthermore, Inoue et al.~\cite{inoue2018transfer} employ transfer learning techniques to mitigate the domain disparity between synthetic and real domain. By leveraging a substantial dataset of synthetic images processed through variational auto-encoders, their approach consistently yields strong performance across diverse conditions.

Constructing simulation environments has been identified as another major challenge, as the settings employed in the aforementioned studies are often not publicly available, which makes it difficult to achieve widespread applicability. Efforts have been made towards developing a versatile simulator to address this issue, facilitating broader generalization and reproducibility~\cite{denninger2019blenderproc, morrical2021nvisii, borkman2021unity}. For example, BlenderProc~\cite{denninger2019blenderproc}, which is built on Blender~\cite{blender2019}, provides a flexible procedural framework capable of producing photorealistic scene images, alongside their respective segmentation masks, depth maps, or normal images. NViSII~\cite{morrical2021nvisii} presents a Python-based rendering tool leveraging NVIDIA's OptiX ray tracing engine and its AI denoiser. This tool is designed for the creation of high-quality synthetic images, accompanied by diverse metadata types for various computer vision tasks, including bounding boxes, segmentation masks, depth maps, normal maps, material properties, and optical flow vectors. Moreover, Unity~\cite{borkman2021unity} advances these initiatives by introducing the Unity Perception package, designed to streamline and expedite synthetic dataset generation. This package provides a user-friendly and highly customizable suite of tools, enabling the production of perfectly annotated data samples. It also features a comprehensive randomization framework to introduce diversity into the datasets created. However, these simulators fall short of integrating with specific domains or tasks, particularly when dealing with objects that have inconsistent shapes and indistinct edges.

In contrast, our approach leverages 3D model of digital twins as well as spacial models of fictional spaces as the simulation framework. A digital twin is a virtual replica of a physical object, process, or system, bridging the physical and digital worlds through real-time data collection and analysis~\cite{grieves2017digital}. This innovative concept has found effective applications across numerous sectors, notably in the manufacturing industry~\cite{kritzinger2018digital}, healthcare~\cite{9255249}, and smart cities~\cite{white2021digital}. To the best of our knowledge, we are the first to employ digital twins for producing synthetic data for fuzzy object detection, resulting in models that surpass those trained on real datasets in performance.

\section{Methodology}
\label{methoddology}
The present method involves utilizing a fully synthetic 3D model of environments and spaces in conjunction with synthetically generated fire 3D models, and an algorithm for automated annotation to create fully synthetic training data for computer vision machine learning (ML) algorithms. 
We utilized EmberGen software, to create fire and flames in visible light spectrum for the synthetic data generation (Figure \ref{fig:obnecol}). Moreover, multiple 3D models created based on real world or fictional environments were utilized to generate the frames that then were fused with synthetically generated fire and flame data. The resulting fully synthetic images are then used for ML model training. The spacial models used to generate the synthetic image backgrounds were collected from the open 3D model libraries of Sketchfab on the web.

 We conducted this study in two subsequent iterations; In Method 1, we utilized four 3D models extracted from real-world digital twins as well as fictional worlds as the backgrounds. The background models were used as a carrier of the synthetic fire frames while randomizing the placement of the fire frames inside the 3D model and only keeping the orientation of the fires constant. We also developed a first version of a code for annotation automation. Subsequently, in Method 1, 1000 fully synthetic images were generate and used to train a CV ML model in Experiment 1. Afterwards, the model performance in fire detection was evaluated to identify improvements. After the analysis of the results of the Method 1, in the subsequent step, efforts were focused on the improvements in the synthetic frames diversity, and annotation accuracy. In Method 2, the diversity of fires and spacial models were significantly increased by using larger number of 3D models and also rendering many different fire and flames separately. Moreover, the annotation accuracy was improved by revising the initial code with a new approach.  

\begin{figure}[t]
  \centering
  
   \includegraphics[width=0.6\linewidth]{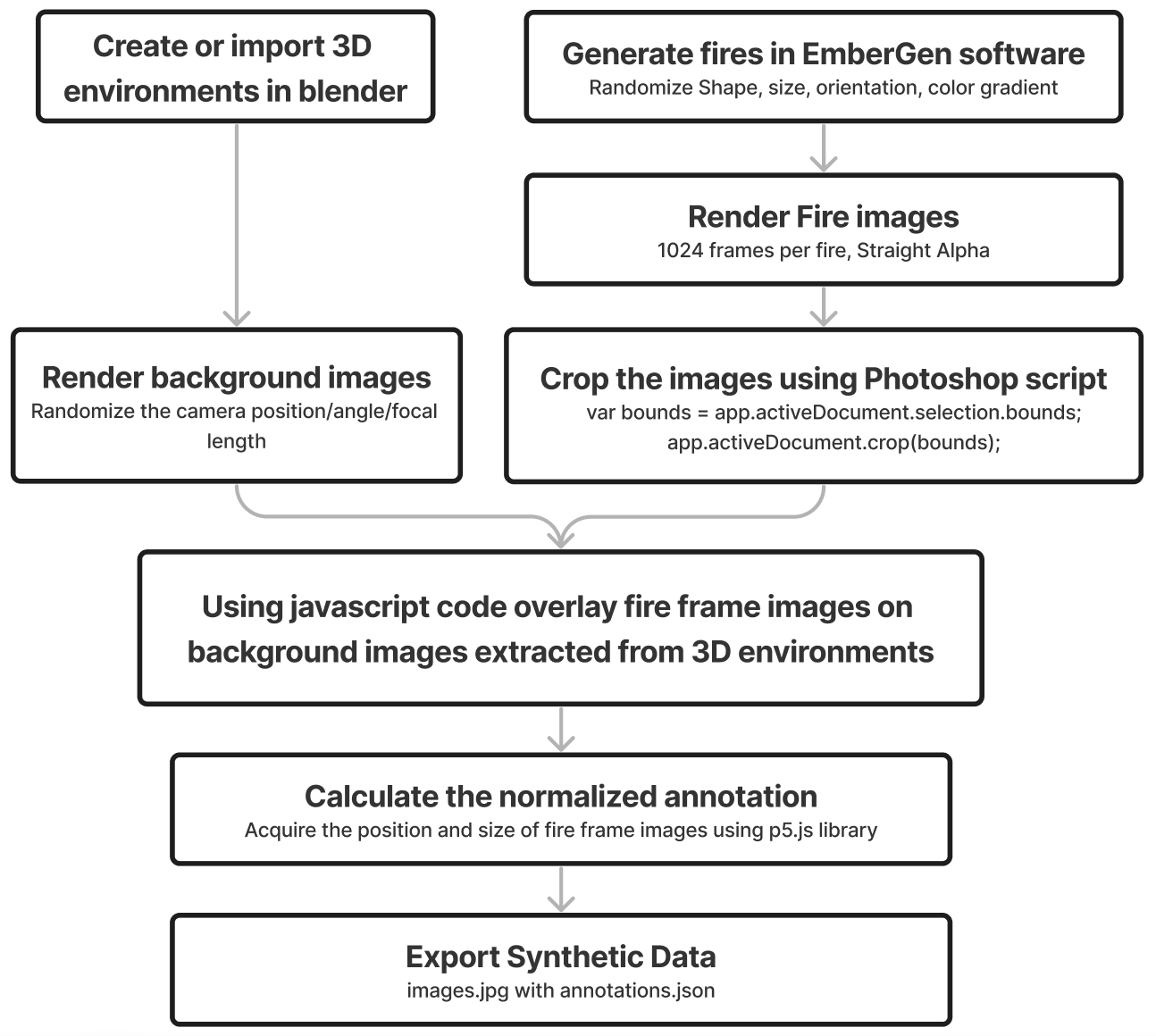}

   \caption{ The Method 2 flowchart for generating and automatically annotating the fully synthetic fire imagery}
   \label{fig:onecol}
\end{figure}

 The details of the improved methodology for the fully synthetic data generation used for machine learning algorithm training for fire detection is depicted in Figure \ref{fig:onecol}.
 
\section{Iterations of synthetic data generation method}
\label{iterations}
To answer the research questions and examine the viability of our proposed synthetic generation method, we conducted two consecutive iterations. In this section, initially, we explain the method proposed for synthetic data generation and automated annotation in Method 1 and complementary Method 2.


\subsection{Method 1: Process establishment}
\label{method1}

\subsubsection{Setting up virtual scenes using 3D models }
\label{method1.DT}
As a first step for our method, we selected 3D scenes reconstructed based on field scans, as well as some open-source scenes from Sketchfab, including residential, commercial, and industrial spaces.

\subsubsection{Defining camera movement rules}
\label{method1.cam}
After selecting the 3D environments, we defined the movement limitations of the camera within each scene, including the area within which the camera should move, rotation angles, and track-to targets. This enables the camera to cover as many different corners of the scene as possible while maintaining a horizontal view, thereby generating a variety of environmental visualizations.

\subsubsection{Generating and arranging flame/fire image planes}
\label{method1.planes}
In this step, the fire and flame images are randomly added to the 3D environmental models and further adjustments are performed. Based on the camera's position, we arranged the flame image planes. Here, flame images are used instead of particle-generated flames because using image flames not only significantly reduces performance overhead but also allows for the use of different flame textures in each frame. The flame images are sourced from EmberGen-generated photo-realistic flame model renders. By adjusting different parameters, various colors, burning shapes, and smoke states the flames can be created and differentiated.

\begin{figure}[t]
  \centering
  
   \includegraphics[clip, trim=1cm 5cm 1cm 7cm, width=1\linewidth]{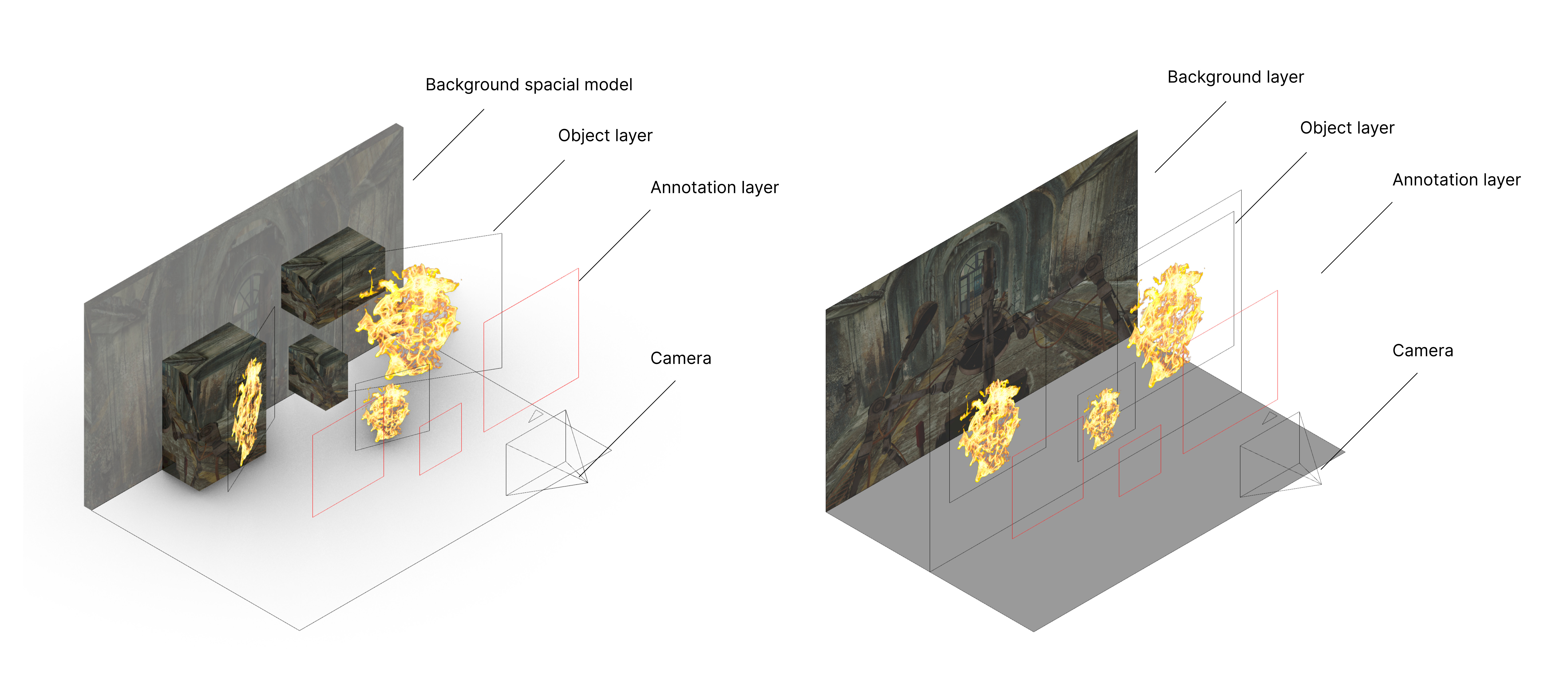}

   \caption{Comparison of background layers, flame object layers, annotation layers, and their relation to camera position used in Method 1 and Method 2. Method 1 (left): Random fire image plane overlay. Method 2 (right): Parallel with background overlay of fire image plane. }
   \label{fig:onecolss}
\end{figure}

\begin{figure*}[t]
  \centering
  
   \includegraphics[width=1\linewidth]{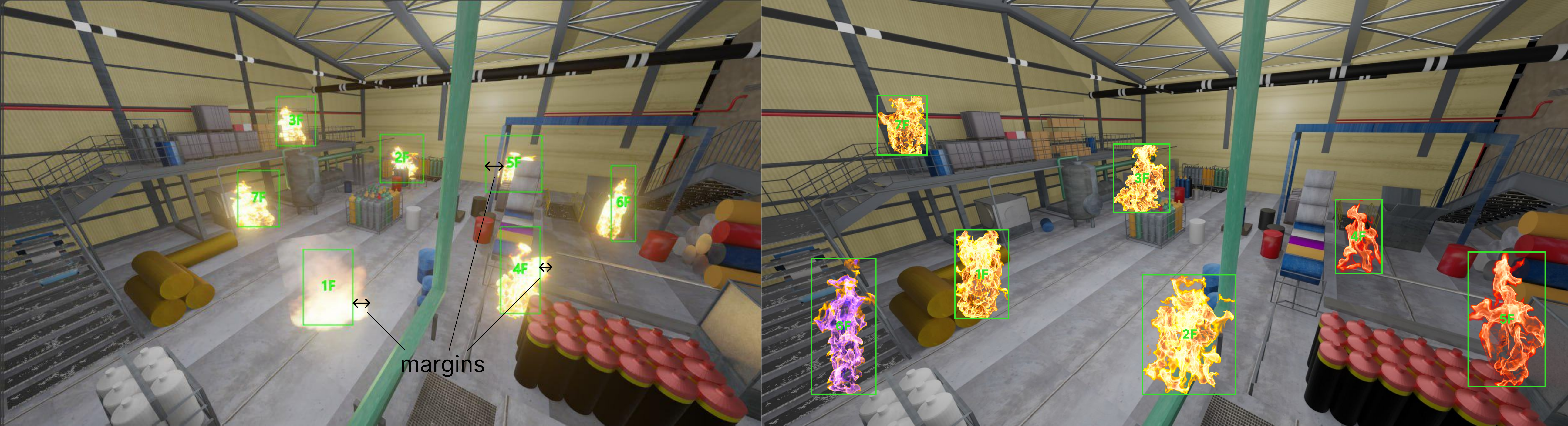}

   \caption{Comparison of annotation bounding box accuracy between 3D flame plane placement in Method 1 (left) and 2D layer flame generation in Method 2 (right) }
   \label{fig:on44ecol}
\end{figure*}

\subsubsection{Calculating annotations}
\label{method1.annot}
After setting up the camera and fires, a code was written in Blender Python to calculate the flame and fire annotations. The logic of the code is to obtain the world coordinates of the flame and fires in the camera plane and use perspective algorithms to translate them into two-dimensional coordinates within the camera plane. On this basis, we utilized the bounding box algorithm and the size of the flame and fire planes to calculate the size of the flame and fires in the camera plane. We output this size and position as a rectangle's center and dimensions into the frame's corresponding JSON file. In cases where the flame or fire are at the edge of the screen, We only calculated the visible part that enters the camera plane (Figure \ref{fig:short1}).

\begin{figure}[t]
  \centering
  
   \includegraphics[width=1\linewidth]{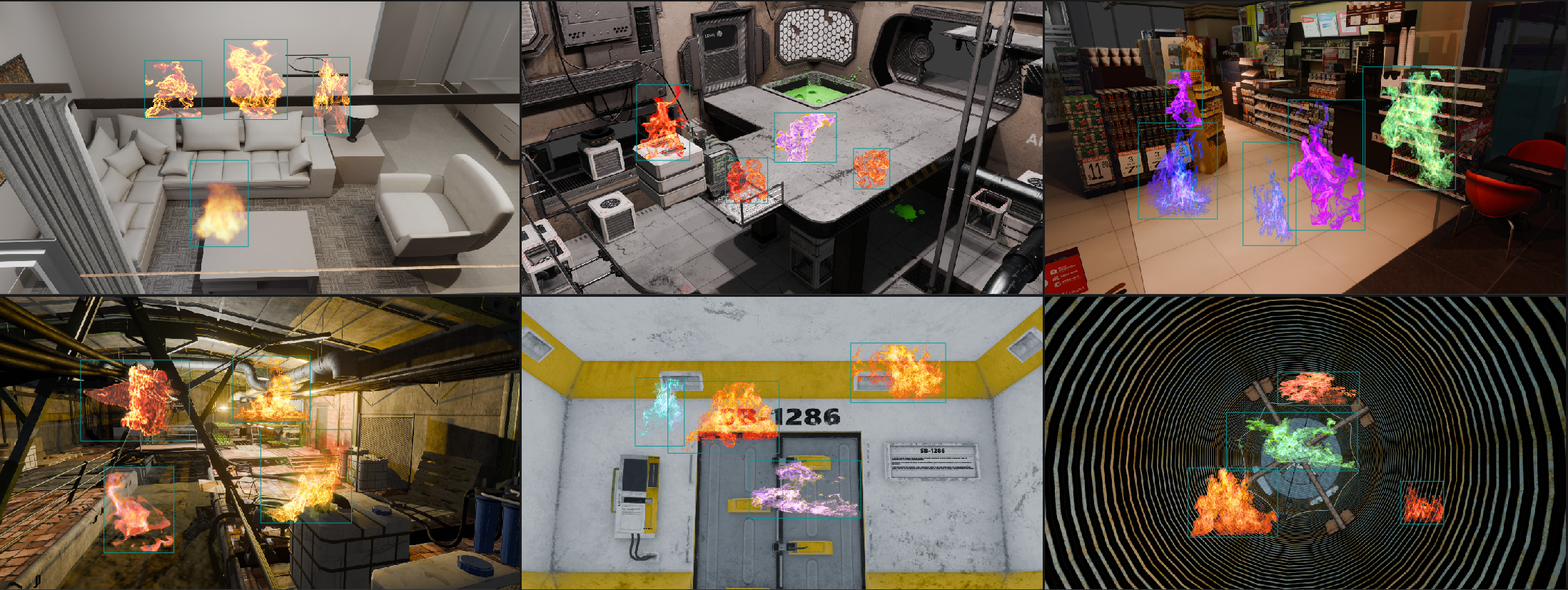}

   \caption{Samples of SynthData 2 (synthetic images generated and annotated in Method 2) used in Experiment 2}
   \label{fig:synth_smpl}
\end{figure}

\subsection{Method 2: Process improvements}
\label{method2}

After utilizing the synthetic data generated in Method 1 for the training of a CV ML model, relatively weak results were achieved, which indicated two major issues in the first method:

\begin{enumerate}
    \item Inaccuracy in the automatic annotation algorithm,
    \item Lack of diversity and randomization in the 3D scenes and the fire and flames generated by EmberGen software.

\end{enumerate}

In Method 2, we addressed the issues, which are explained as follows;
\subsubsection{Arranging flame/fire image planes to improve annotations}
\label{method2.planes}
Step \ref{method1.planes} of Method 1, caused challenges with the accuracy of the automated annotation of the synthetic fires and flames in the synthetic scenes (Figure \ref{fig:short1}). To resolve this issue in the Method 2, we imported the EmberGen-generated flame images into the Blender scene in the form of 'image to plane’ and set these planes' constraints to 'always face the camera' (Figure \ref{fig:onecolss}). Subsequently, we randomized the positions and sizes of these flame planes according to the camera's location.

The details of the improvements made to the annotation process in Method 2 are outlined as follows;

To enhance the precision of flame annotation within 3D scenes, an alternative approach was adopted that involved directly overlaying fire and flame images of varying sizes and positions onto a pre-rendered background image. The process of defining camera movement rules explained in section \ref{method1.cam} was repeated, but this time the 3D scene did not contain preset fire and flames. After obtaining the same rendered images of the 3D scene, we extracted these images and used them as the background for the next step. In the subsequent step, fire and flame frames were randomly placed in the foreground. This method consolidates the steps of placing fire and flame planes within a 3D environment and projecting them onto the camera plane from the previous method. In this way, as the flames were directly placed on the 2D plane, we could achieve very accurate annotations and effectively eliminate errors caused by perspective distortion due to camera focus (Figure \ref{fig:on44ecol}).

\subsubsection{Diversification of the scenes and fires}
\label{method2.divers}

Additionally, we improved the logic for generating fire and flames with EmberGen, allowing us to produce a rich variety of flame shapes and colors. The new fire and flame images were the result of rendering a fire 3D model with varying criteria and then converting the video of the burning fire into 2D frames. We selected flame images from a continuous set of 1024 image files, picking one every 12 frames to ensure the distinctiveness of the flame shapes (Figure \ref{fig:onecol66}). Criteria changed for each fire and flame model were particle emission method, emission shape, and color. Thus, we could easily obtain thousands of different fire and flame images. After rendering the fire and flame models and obtaining the frames, we used a Photoshop script for batch processing to remove the transparent background and crop to the size of the colored pixel part, to ensure the accuracy of the annotations.

In addition to fire and flame diversification, we also increased the number of 3D environmental models used for generating the scene backgrounds. While in Method 1, we only used 4 environmental models, in Method 2, 48 environments and distinct scenes were utilized assisting with the improved diversity and randomization of the synthetic imagery (Figure \ref{fig:synth_smpl}). 

\section{ML experiments }
\label{exp}
 \label{exp}
To examine the viability of our fuzzy object synthetic data generation methods, we conducted a series of ML experiments.
Our experiments were inspired by previous research \cite{hinterstoisser2019annotation}, \cite{borkman2021unity}.
We describe our experimental setup in Section \ref{exp.setup}, followed by a comparison of the performance results of our two synthetic image generation methods in Section \ref{exp.cmp_method}. Finally, we compare the performance of different training combinations of Experiment 2 on two different test benchmarks in Section \ref{exp.cmp_cmbs}.

\subsection{Experimental setup}
\label{exp.setup}

\textbf{Datasets.} We introduce two synthetic fire datasets: SynthData 1 and SynthData 2, each containing 1,000 images and 2D bounding box annotations generated using Method 1 (\ref{method1}) and Method 2 (\ref{method2}) respectively. We compare the performance of these two synthetic fire datasets (Experiment 1 vs. Experiment 2) in Section \ref{exp.cmp_method}.
Experiment 1 is training of YOLOv5 mini using SynthData 1 while Experiment 2 is the training of YOLOv5 mini using SynthData 2.

We also collected a real-world diverse dataset, named "Fire in The Wild" FiIW, containing 1400 images of real instances of fire scenes. To curate this dicverse dataset, we first extended the D-Fire \cite{de2022automatic} dataset with more web-source fire images, then removed all exact-duplicate and near-duplicate images, based on the Euclidean distance of their VGG embeddings, and finally selected the top 1,400 unique images, based on the ranking of their relative distance to other samples. This dataset is annotated manually by us and with the guidelines from the VOC2011 \cite{voc2011_annot_guidline}. It is randomly shuffled and split into a training set of 1000 (71.4\%) images, a validation set of 200 (14.3\%) images, and a test set of 200 (14.3\%) images. These validation and test sets are used in both experiments.

We also randomly selected a few subsets of size 250 (25\%), 500 (50\%), and 750 (75\%) from both the FiIW training set and the SynthData 2, to build multiple mixtures of synthetic and real images for training in Section \ref{exp.cmp_cmbs}. The total number of images in each training mixture is kept constant 1,000.

To assess the generalizability of the model in identifying rare instances of fires, an additional real-world web-source test dataset was curated, denoted as "RealRareFire", which also is manually annotated by us and with the guidelines from the VOC2011 \cite{voc2011_annot_guidline}. This dataset comprises 69 images depicting uncommon fire occurrences, distinct from those encompassed within the FiIW dataset. Although RealRareFire is relatively small as a dataset, it is noteworthy that the acquisition of each image in that demanded considerably more time compared to that of the FiIW dataset. This is attributable to the scarcity of such images exhibiting distinctive visual attributes, namely color and shape, across online sources. The comparative evaluation of the model's performance across these two test datasets is instrumental in discerning potential instances of overfitting to the FiIW dataset, which predominantly comprises fire images readily accessible on the internet.

\textbf{Evaluation Metrics.}
We consider two evaluation metrics of mean Average Precision averaged across IoU thresholds of [0.5:0.95] ($\mathbf{AP}$), and mean Average Precision with a single intersection of union (IoU) threshold of 0.5 ($\mathbf{AP_{50}}$). These are based on MS COCO \cite{lin2015microsoft} definition, which are extensively used across the object detection literature \cite{borkman2021unity,bochkovskiy2020yolov4, wang2022yolov7, ge2021yolox}.

\textbf{Training Setup.} 
In both experiments, we perform the default data augmentation methods in YOLOv5 \cite{jocher_ultralyticsyolov5_2022}, including Mosaic introduced by \cite{bochkovskiy2020yolov4}, image translation, horizontal flipping, HSV (hue saturation value) augmentation. All models are trained with a minibatch size of 64 on one NVIDIA-A100 GPU for 100 epochs with SGD (stochastic gradient descent) for optimization. Other default hyper-parameters in YOLOv5 \cite{jocher_ultralyticsyolov5_2022} are unchanged, including "fitness", which is the metric to select the best model over training epochs, and is computed on the validation set of the FiIW. \cite{jocher_ultralyticsyolov5_2022} defined it as: 
\[Fitness = 0.1 \times AP_{50} + 0.9 \times AP\]
We decided to use the same coefficient weights for $\mathbf{AP_{50}}$ and $\mathbf{AP}$ to facilitate the reproducibility of our work, although YOLOv5 \cite{jocher_ultralyticsyolov5_2022} lacks a clear explanation for choosing these weights.

\textbf{Random Seeds.} As there is inherent randomness in choosing the subsets and in training neural networks, we report the mean and standard deviation of each metric over five seeds.

\subsection{Comparison of synthetic data generation methods}
\label{exp.cmp_method}

\begin{table}[b]
\centering

\caption{Comparison of Experiments 1 and 2 mean and standard deviation over five seeds}

\label{table:cmp_method}
\footnotesize 
\begin{tabular}{@{}lcccc@{}}
\toprule
& \multicolumn{2}{c}{\textbf{$\mathbf{FiIW_{test}}$}} & \multicolumn{2}{c}{\textbf{$\mathbf{RealRareFire}$}} \\
\cmidrule(lr){2-3} \cmidrule(lr){4-5}
& $\mathbf{AP_{50} (\%)}$ & $\mathbf{AP (\%)}$ & $\mathbf{AP_{50} (\%)}$ & $\mathbf{AP (\%)}$ \\

\midrule
Experiment 1 & 3.64 $\pm$ 1.26 & 0.94 $\pm$ 0.37 & 4.9 $\pm$ 1.14 & 1.35 $\pm$ 0.33\\
Experiment 2 & \textbf{15.14 $\pm$ 1.16} & \textbf{5.65 $\pm$ 0.61}   & \textbf{24.38 $\pm$ 2.09} & \textbf{8.87 $\pm$ 0.52} \\

\bottomrule
\end{tabular}
\end{table}
\normalsize 

\begin{figure*}[t]
  \centering
  \begin{subfigure}{0.49\linewidth}
    \includegraphics[width=\linewidth]{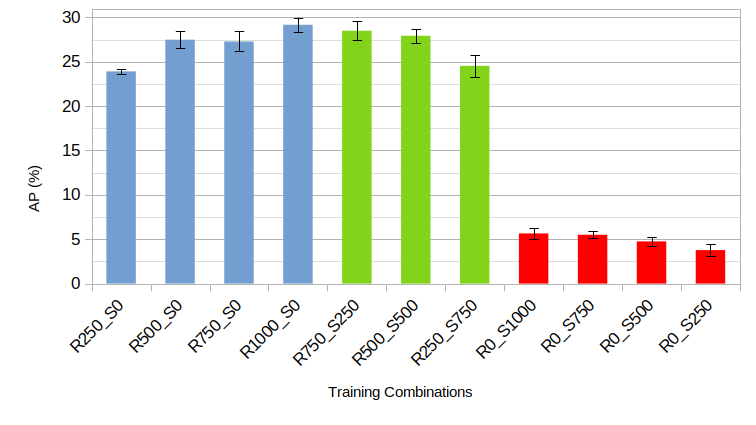}
    \caption{}
  \end{subfigure}
  \hfill
  \begin{subfigure}{0.49\linewidth}
    \includegraphics[width=\linewidth]{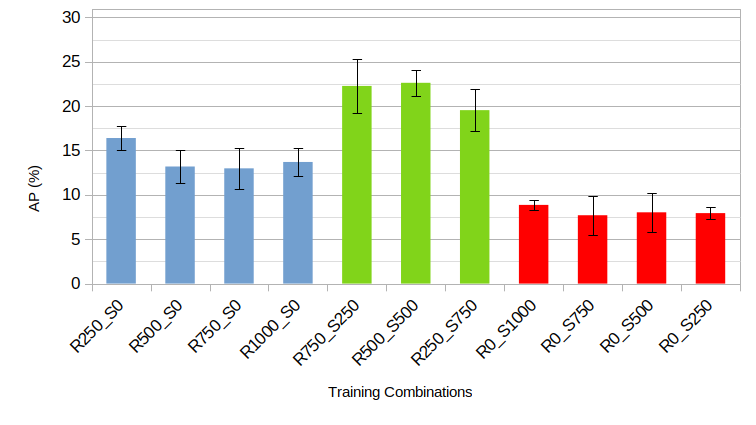}
    \caption{}
  \end{subfigure}
  \caption{Comparison of training combinations tested on (a) $\mathbf{FiIW_{test}}$ and (b) $\mathbf{RealRareFire}$}
  \label{fig:result}
\end{figure*}

Table \ref{table:cmp_method} shows the detection performance result obtained when using different iterations in our method. Experiment 1 uses synthetic data generated and annotated using Method 1 with low level of scene and fire randomization (SynthData 1). Experiment 2 uses synthetic data generated and annotated using Method 2 with high level of scene and fire randomization (SynthData 2). 

In addressing our initial research question, it becomes evident that ML models trained on both iterations of the fully synthetic fire data generation method produce fire detection. However, we observe that the SynthData 2 outperforms SynthData 1 based on the results reported in Table \ref{table:cmp_method}. Therefore, Method 2 outperforms Method 1 by high margins in all metrics and benchmarks. This relates to the challenges identified in the Method 1 and addressed in Method 2 (see \ref{method2})

\subsection{Comparison of various training combinations}
\label{exp.cmp_cmbs}
\textbf{Combination Setup.}  Three training strategies with different combinations of SynthData 2 and real-world data are used in this part:
\begin{itemize}
    \item In Strategy 1, four subsets (250, 500, 750, 1000) of the training set of FiIW is used to fine-tune the model.
    
    \item In Strategy 2, we control for the size of training dataset to be 1000, and build three mixtures of real-synthetic training sets using (250, 500, 750) subsets of each real and synthetic dataset.
    
    \item In Strategy 3, we use four subsets (250, 500, 750, 1000) of SynthData 2 for ML model training.
\end{itemize}

\begin{table*}[ht]
\centering

\caption{Comparison of three strategies and  their training set combinations in R(m)\_S(n) format, where m and n indicate the number of images from Real training and Synthetic sets respectively. Comparison metrics are $AP_{50}$ and $AP$ (the higher, the better). Mean and standard deviation over 5 seeds is shown for all combinations. \textbf{Best} results are shown in \textbf{bold}, \textit{second best} results are shown in \textit{italic}.}

\label{table:result}
\small 
\begin{tabular}{@{}llcccc@{}}
\toprule
& & \multicolumn{2}{c}{\textbf{$\mathbf{FiIW_{test}}$}} & \multicolumn{2}{c}{\textbf{$\mathbf{RealRareFire}$}} \\
\cmidrule(lr){3-4} \cmidrule(lr){5-6}
& & $\mathbf{AP_{50} (\%)}$ & $\mathbf{AP (\%)}$ & $\mathbf{AP_{50} (\%)}$ & $\mathbf{AP (\%)}$ \\

\midrule
\multirow{4}{*}{\textcolor{CadetBlue}{Strategy 1}} & \textbf{R250\_S0}   & 51.68 $\pm$ 1.11 & 23.92 $\pm$ 0.31  & 35.41 $\pm$ 2.57 & 16.41 $\pm$ 1.3 \\

& \textbf{R500\_S0}   & 56.05 $\pm$ 0.82 & 27.49 $\pm$ 0.95  & 26.69 $\pm$ 5.25 & 13.2 $\pm$ 1.91 \\

& \textbf{R750\_S0}   & 55.51 $\pm$ 1.22 & 27.29 $\pm$ 1.13 & 25.37 $\pm$ 4.32 & 12.99 $\pm$ 2.34 \\

& \textbf{R1000\_S0}  & \textbf{58.13 $\pm$ 1.05} & \textbf{29.17 $\pm$ 0.79}  & 26.18 $\pm$ 2.61 & 13.71 $\pm$ 1.6 \\

\midrule

\multirow{3}{*}{\textcolor{LimeGreen}{Strategy 2}} & \textbf{R750\_S250} & \textit{56.8 $\pm$ 1.39} & \textit{28.5 $\pm$ 1.05}  & \textit{41.54 $\pm$ 5.24} & \textit{22.28 $\pm$ 3.08} \\

& \textbf{R500\_S500} & 56.74 $\pm$ 1.38  & 27.93 $\pm$ 0.8  & \textbf{42.3 $\pm$ 3.62}  & \textbf{22.63 $\pm$ 1.49} \\

& \textbf{R250\_S750} & 52.37 $\pm$ 1.47 & 24.54 $\pm$ 1.21  & 39.77 $\pm$ 3.92 & 19.55 $\pm$ 2.35 \\

\midrule
\multirow{4}{*}{\textcolor{red}{Strategy 3}} & \textbf{R0\_S1000}  & 15.14 $\pm$ 1.16 & 5.65 $\pm$ 0.61   & 24.38 $\pm$ 2.09 & 8.87 $\pm$ 0.52 \\
& \textbf{R0\_S750}   & 14.43 $\pm$ 1.61 & 5.5 $\pm$ 0.38     & 20.01 $\pm$ 5.32 & 7.7 $\pm$ 2.2 \\
& \textbf{R0\_S500}   & 12.87 $\pm$ 1.57 & 4.75 $\pm$ 0.55   & 22.18 $\pm$ 7.03 & 8.03 $\pm$ 2.22 \\
& \textbf{R0\_S250}   & 10.56 $\pm$ 1.96 & 3.77 $\pm$ 0.72   & 21.99 $\pm$ 2.45 & 7.94 $\pm$ 0.72 \\
\bottomrule
\end{tabular}
\end{table*}

\normalsize 

\textbf{Findings.} Table \ref{table:result} and Figure \ref{fig:result} present the detection performance of different combinations of training set on the two test sets. The findings of the these experiments are as follows:
\begin{itemize}
    \item In Strategy 1, we observe that with more real data, the model performance is improved under FiIW setup, and diminished in the RealRareFire setup. This indicates that increasing the number of real data in a purely real training set harms model generalization to real rare fire cases. We also observe that all models in this strategy perform better under FiIW setting, for example, R1000\_S0 has $\sim$15.5\% lower AP under RealRareFire setting. This indicates that the models are not able to generalize to real rare fire cases in terms of color and shape, and are over-fitted to common fire cases.
    
    \item In Strategy 2, R750\_S250 and R500\_S500 show approximately similar performance while both having higher performance than R250\_S750 under both test sets. Nevertheless, R500\_S500 having the best performance overall under RealRareFire test set, and, unlike Strategy 1, adding more real data to a mixed real-synthetic training set does not harm the performance on RealRareFire, which is another benefit of using both types of data. Furthermore, we observe a smaller gap in range of $\sim$[5,6]\% between the AP score under both test sets.
    
    \item In Strategy 3, we see that with more synthetic data, the model performance increases under FiIW setup; while, having no substantial improvement under RealRareFire setup. Unlike Strategy 1, all four models in Strategy 3 do not show substantial performance improvements under RealRareFire setting. For instance, R0\_S1000 has only $\sim$3\% higher AP under RealRareFire setting. This indicates that increasing the number of fully synthetic data in a purely synthetic training set does not have substantial effect on model generalization to real rare fire cases.
    
    \item Comparing all strategies, first, we observe that all three combinations of Strategy 2 outperform the Strategies 1 and 3 under RealRareFire test set. Secondly, although R1000\_S0 from Strategy 1 performs the best under FiIW test set, Strategy 2 is on a par with that, where R750\_S250 and R500\_S500 are the \textbf{second} and the \textbf{third best} overall performers respectively, while potentially having \textbf{lower cost} in terms of collection and annotation.
\end{itemize}
It is worth noting that, similar to Borkman et al. \cite{borkman2021unity}, since all experiments are conducted with one model and a limited training size (1000 images or smaller). Therefore, the experimental results might not generalize to other models or training dataset sizes.

\section{Conclusion}

The contribution of this research is to overcome the shortcomings of the present fuzzy object computer vision training methods by proposing a synthetic data generation and annotation method and to disclose its comparative performance and shortcomings. Our proposed method showed promising results for the detection of fire in both common and rare fire test sets. Moreover, we illustrated that a mixed model trained on a fully synthetic and real training sets can outperform a model only trained on real data and better generalize in detection of real fire instances with various visual features. We clearly illustrated that, randomization and diversity of the synthetic data is necessary for the method to work, while only generating realistic imagery alone is not effective for generalization. The limitations of this research was the use of limited number of synthetic images that should be increased in the future research.

\subsection{Future research}

The lack of large datasets is only one of the reasons the current CNN based method performs poorly on fuzzy object detection. Most of the popular detection models, such as YOLO~\cite{redmon2016you}, ViT~\cite{dosovitskiy2020image}, focus on detecting static objects like cats, dogs, and people, which have fixed shape and the shape will not change over time. However, fuzzy objects have variable shapes and move over time, which are different from static objects and these dynamic characteristics cannot be described in a single image. Therefore, it is harder for the current CNN based method to capture the features of these types of objects. However, a model trained directly on video may be capable of detecting temporal patterns of shape shift and movement behaviors of fuzzy objects in a video, which are not available in an static image. Therefore, training on video may be a better choice than training on image for fuzzy objects. In literature, video classification methods have been widely used to analyze human actions~\cite{carreira2017quo},~\cite{carreira2018short},~\cite{bertasius2021space}. However, to our knowledge, there is no research about training video classification models for fuzzy objects and the use of synthetic videos for this purpose.
Moreover, future research should investigate the use of our proposed method for other fuzzy objects such as smoke. This should be studied in relation to the sensitivity, confidence level, and accuracy of a computer vision model trained on real smoke imagery.
The future research may also study the use of custom synthetic data generated in environmental digital twins for custom retraining of the generic fuzzy object detection models for improved performance.

\bibliographystyle{unsrtnat}
\bibliography{main}

\begin{thebibliography}{41}
\providecommand{\natexlab}[1]{#1}
\providecommand{\url}[1]{\texttt{#1}}
\expandafter\ifx\csname urlstyle\endcsname\relax
  \providecommand{\doi}[1]{doi: #1}\else
  \providecommand{\doi}{doi: \begingroup \urlstyle{rm}\Url}\fi

\bibitem[Krizhevsky et~al.(2012)Krizhevsky, Sutskever, and Hinton]{krizhevsky2012imagenet}
Alex Krizhevsky, Ilya Sutskever, and Geoffrey~E Hinton.
\newblock Imagenet classification with deep convolutional neural networks.
\newblock \emph{Advances in neural information processing systems}, 25, 2012.

\bibitem[Deng et~al.(2009)Deng, Dong, Socher, Li, Li, and Fei-Fei]{deng2009imagenet}
Jia Deng, Wei Dong, Richard Socher, Li-Jia Li, Kai Li, and Li~Fei-Fei.
\newblock Imagenet: A large-scale hierarchical image database.
\newblock In \emph{2009 IEEE conference on computer vision and pattern recognition}, pages 248--255. Ieee, 2009.

\bibitem[Zhang et~al.(2016)Zhang, Xu, Xu, and Guo]{zhang2016deep}
Q.~Zhang, J.~Xu, L.~Xu, and H.~Guo.
\newblock Deep convolutional neural networks for forest fire detection.
\newblock In \emph{2016 International Forum on Management, Education and Information Technology Application}. Atlantis Press, 2016.

\bibitem[Muhammad et~al.(2018{\natexlab{a}})Muhammad, Ahmad, and Baik]{muhammad2018early}
K.~Muhammad, J.~Ahmad, and S.~W. Baik.
\newblock Early fire detection using convolutional neural networks during surveillance for effective disaster management.
\newblock \emph{Neurocomputing}, 288:\penalty0 30--42, 2018{\natexlab{a}}.

\bibitem[Muhammad et~al.(2018{\natexlab{b}})Muhammad, Ahmad, Lv, Bellavista, Yang, and Baik]{muhammad2018efficient}
K.~Muhammad, J.~Ahmad, Z.~Lv, P.~Bellavista, P.~Yang, and S.~W. Baik.
\newblock Efficient deep cnn-based fire detection and localization in video surveillance applications.
\newblock \emph{IEEE Transactions on Systems, Man, and Cybernetics: Systems}, \penalty0 (99):\penalty0 1--16, 2018{\natexlab{b}}.

\bibitem[Muhammad et~al.(2018{\natexlab{c}})Muhammad, Ahmad, Mehmood, Rho, and Baik]{muhammad2018convolutional}
K.~Muhammad, J.~Ahmad, I.~Mehmood, S.~Rho, and S.~W. Baik.
\newblock Convolutional neural networks based fire detection in surveillance videos.
\newblock \emph{IEEE Access}, 6:\penalty0 18174--18183, 2018{\natexlab{c}}.

\bibitem[Muhammad et~al.(2019)Muhammad, Khan, Elhoseny, Ahmed, and Baik]{muhammad2019efficient}
K.~Muhammad, S.~Khan, M.~Elhoseny, S.~H. Ahmed, and S.~W. Baik.
\newblock Efficient fire detection for uncertain surveillance environment.
\newblock \emph{IEEE Transactions on Industrial Informatics}, 2019.

\bibitem[Foggia et~al.(2015)Foggia, Saggese, and Vento]{foggia2015real}
P.~Foggia, A.~Saggese, and M.~Vento.
\newblock Real-time fire detection for videosurveillance applications using a combination of experts based on color, shape, and motion.
\newblock \emph{IEEE TRANSACTIONS on circuits and systems for video technology}, 25\penalty0 (9):\penalty0 1545--1556, 2015.

\bibitem[de~Venancio et~al.(2022)de~Venancio, Lisboa, and Barbosa]{de2022automatic}
Pedro Vinicius~AB de~Venancio, Adriano~C Lisboa, and Adriano~V Barbosa.
\newblock An automatic fire detection system based on deep convolutional neural networks for low-power, resource-constrained devices.
\newblock \emph{Neural Computing and Applications}, 34\penalty0 (18):\penalty0 15349--15368, 2022.

\bibitem[Simonyan and Zisserman(2015)]{simonyan2015vgg}
Karen Simonyan and Andrew Zisserman.
\newblock Very deep convolutional networks for large-scale image recognition, 2015.

\bibitem[Tobin et~al.(2017)]{tobin2017domain}
Josh Tobin et~al.
\newblock Domain randomization for transferring deep neural networks from simulation to the real world.
\newblock In \emph{2017 IEEE/RSJ International Conference on Intelligent Robots and Systems (IROS)}. IEEE, 2017.

\bibitem[Hinterstoisser et~al.(2019)]{hinterstoisser2019annotation}
Stefan Hinterstoisser et~al.
\newblock An annotation saved is an annotation earned: Using fully synthetic training for object instance detection.
\newblock \emph{arXiv preprint arXiv:1902.09967}, 2019.

\bibitem[Borkman et~al.(2021)Borkman, Crespi, Dhakad, Ganguly, Hogins, Jhang, Kamalzadeh, Li, Leal, Parisi, Romero, Smith, Thaman, Warren, and Yadav]{borkman2021unity}
Steve Borkman, Adam Crespi, Saurav Dhakad, Sujoy Ganguly, Jonathan Hogins, You-Cyuan Jhang, Mohsen Kamalzadeh, Bowen Li, Steven Leal, Pete Parisi, Cesar Romero, Wesley Smith, Alex Thaman, Samuel Warren, and Nupur Yadav.
\newblock Unity perception: Generate synthetic data for computer vision, 2021.

\bibitem[Everingham et~al.(2010)Everingham, Van~Gool, Williams, Winn, and Zisserman]{everingham2010pascal}
Mark Everingham, Luc Van~Gool, Christopher~KI Williams, John Winn, and Andrew Zisserman.
\newblock The pascal visual object classes (voc) challenge.
\newblock \emph{International journal of computer vision}, 88:\penalty0 303--338, 2010.

\bibitem[Lin et~al.(2015)Lin, Maire, Belongie, Bourdev, Girshick, Hays, Perona, Ramanan, Zitnick, and Dollár]{lin2015microsoft}
Tsung-Yi Lin, Michael Maire, Serge Belongie, Lubomir Bourdev, Ross Girshick, James Hays, Pietro Perona, Deva Ramanan, C.~Lawrence Zitnick, and Piotr Dollár.
\newblock Microsoft coco: Common objects in context, 2015.

\bibitem[He et~al.(2016)He, Zhang, Ren, and Sun]{he2016deep}
Kaiming He, Xiangyu Zhang, Shaoqing Ren, and Jian Sun.
\newblock Deep residual learning for image recognition.
\newblock In \emph{Proceedings of the IEEE conference on computer vision and pattern recognition}, pages 770--778, 2016.

\bibitem[Ren et~al.(2015)Ren, He, Girshick, and Sun]{ren2015faster}
Shaoqing Ren, Kaiming He, Ross Girshick, and Jian Sun.
\newblock Faster r-cnn: Towards real-time object detection with region proposal networks.
\newblock \emph{Advances in neural information processing systems}, 28, 2015.

\bibitem[Redmon et~al.(2016)]{redmon2016you}
Joseph Redmon et~al.
\newblock You only look once: Unified, real-time object detection.
\newblock In \emph{Proceedings of the IEEE conference on computer vision and pattern recognition}, 2016.

\bibitem[Lin et~al.(2017)Lin, Goyal, Girshick, He, and Doll{\'a}r]{lin2017focal}
Tsung-Yi Lin, Priya Goyal, Ross Girshick, Kaiming He, and Piotr Doll{\'a}r.
\newblock Focal loss for dense object detection.
\newblock In \emph{Proceedings of the IEEE international conference on computer vision}, pages 2980--2988, 2017.

\bibitem[He et~al.(2017)He, Gkioxari, Doll{\'a}r, and Girshick]{he2017mask}
Kaiming He, Georgia Gkioxari, Piotr Doll{\'a}r, and Ross Girshick.
\newblock Mask r-cnn.
\newblock In \emph{Proceedings of the IEEE international conference on computer vision}, pages 2961--2969, 2017.

\bibitem[Long et~al.(2015)Long, Shelhamer, and Darrell]{long2015fully}
Jonathan Long, Evan Shelhamer, and Trevor Darrell.
\newblock Fully convolutional networks for semantic segmentation.
\newblock In \emph{Proceedings of the IEEE conference on computer vision and pattern recognition}, pages 3431--3440, 2015.

\bibitem[Su et~al.(2015)Su, Qi, Li, and Guibas]{su2015render}
Hao Su, Charles~R Qi, Yangyan Li, and Leonidas~J Guibas.
\newblock Render for cnn: Viewpoint estimation in images using cnns trained with rendered 3d model views.
\newblock In \emph{Proceedings of the IEEE international conference on computer vision}, pages 2686--2694, 2015.

\bibitem[Georgakis et~al.(2017)Georgakis, Mousavian, Berg, and Kosecka]{georgakis2017synthesizing}
Georgios Georgakis, Arsalan Mousavian, Alexander~C Berg, and Jana Kosecka.
\newblock Synthesizing training data for object detection in indoor scenes.
\newblock \emph{arXiv preprint arXiv:1702.07836}, 2017.

\bibitem[Shrivastava et~al.(2017)Shrivastava, Pfister, Tuzel, Susskind, Wang, and Webb]{shrivastava2017learning}
Ashish Shrivastava, Tomas Pfister, Oncel Tuzel, Joshua Susskind, Wenda Wang, and Russell Webb.
\newblock Learning from simulated and unsupervised images through adversarial training.
\newblock In \emph{Proceedings of the IEEE conference on computer vision and pattern recognition}, pages 2107--2116, 2017.

\bibitem[Inoue et~al.(2018)Inoue, Choudhury, De~Magistris, and Dasgupta]{inoue2018transfer}
Tadanobu Inoue, Subhajit Choudhury, Giovanni De~Magistris, and Sakyasingha Dasgupta.
\newblock Transfer learning from synthetic to real images using variational autoencoders for precise position detection.
\newblock In \emph{2018 25th IEEE International Conference on Image Processing (ICIP)}, pages 2725--2729. IEEE, 2018.

\bibitem[Denninger et~al.(2019)Denninger, Sundermeyer, Winkelbauer, Zidan, Olefir, Elbadrawy, Lodhi, and Katam]{denninger2019blenderproc}
Maximilian Denninger, Martin Sundermeyer, Dominik Winkelbauer, Youssef Zidan, Dmitry Olefir, Mohamad Elbadrawy, Ahsan Lodhi, and Harinandan Katam.
\newblock Blenderproc.
\newblock \emph{arXiv preprint arXiv:1911.01911}, 2019.

\bibitem[Morrical et~al.(2021)Morrical, Tremblay, Lin, Tyree, Birchfield, Pascucci, and Wald]{morrical2021nvisii}
Nathan Morrical, Jonathan Tremblay, Yunzhi Lin, Stephen Tyree, Stan Birchfield, Valerio Pascucci, and Ingo Wald.
\newblock Nvisii: A scriptable tool for photorealistic image generation.
\newblock \emph{arXiv preprint arXiv:2105.13962}, 2021.

\bibitem[{Blender Online Community}(2019)]{blender2019}
{Blender Online Community}.
\newblock Blender - a 3d modelling and rendering package, 2019.
\newblock [Online]. Available: \url{http://www.blender.org}.

\bibitem[Grieves and Vickers(2017)]{grieves2017digital}
Michael Grieves and John Vickers.
\newblock Digital twin: Mitigating unpredictable, undesirable emergent behavior in complex systems.
\newblock \emph{Transdisciplinary perspectives on complex systems: New findings and approaches}, pages 85--113, 2017.

\bibitem[Kritzinger et~al.(2018)Kritzinger, Karner, Traar, Henjes, and Sihn]{kritzinger2018digital}
Werner Kritzinger, Matthias Karner, Georg Traar, Jan Henjes, and Wilfried Sihn.
\newblock Digital twin in manufacturing: A categorical literature review and classification.
\newblock \emph{Ifac-PapersOnline}, 51\penalty0 (11):\penalty0 1016--1022, 2018.

\bibitem[Erol et~al.(2020)Erol, Mendi, and Doğan]{9255249}
Tolga Erol, Arif~Furkan Mendi, and Dilara Doğan.
\newblock The digital twin revolution in healthcare.
\newblock In \emph{2020 4th International Symposium on Multidisciplinary Studies and Innovative Technologies (ISMSIT)}, pages 1--7, 2020.
\newblock \doi{10.1109/ISMSIT50672.2020.9255249}.

\bibitem[White et~al.(2021)White, Zink, Codec{\'a}, and Clarke]{white2021digital}
Gary White, Anna Zink, Lara Codec{\'a}, and Siobh{\'a}n Clarke.
\newblock A digital twin smart city for citizen feedback.
\newblock \emph{Cities}, 110:\penalty0 103064, 2021.

\bibitem[voc()]{voc2011_annot_guidline}
{VOC2011} {Annotation} {Guidelines}.
\newblock URL \url{http://host.robots.ox.ac.uk/pascal/VOC/voc2012/guidelines.html}.

\bibitem[Bochkovskiy et~al.(2020)Bochkovskiy, Wang, and Liao]{bochkovskiy2020yolov4}
Alexey Bochkovskiy, Chien-Yao Wang, and Hong-Yuan~Mark Liao.
\newblock Yolov4: Optimal speed and accuracy of object detection, 2020.

\bibitem[Wang et~al.(2022)Wang, Bochkovskiy, and Liao]{wang2022yolov7}
Chien-Yao Wang, Alexey Bochkovskiy, and Hong-Yuan~Mark Liao.
\newblock Yolov7: Trainable bag-of-freebies sets new state-of-the-art for real-time object detectors, 2022.

\bibitem[Ge et~al.(2021)Ge, Liu, Wang, Li, and Sun]{ge2021yolox}
Zheng Ge, Songtao Liu, Feng Wang, Zeming Li, and Jian Sun.
\newblock Yolox: Exceeding yolo series in 2021, 2021.

\bibitem[Jocher et~al.(2022)Jocher, Chaurasia, Stoken, Borovec, NanoCode012, Kwon, Michael, TaoXie, Fang, imyhxy, Lorna, Yifu), Wong, V, Montes, Wang, Fati, Nadar, Laughing, UnglvKitDe, Sonck, tkianai, yxNONG, Skalski, Hogan, Nair, Strobel, and Jain]{jocher_ultralyticsyolov5_2022}
Glenn Jocher, Ayush Chaurasia, Alex Stoken, Jirka Borovec, NanoCode012, Yonghye Kwon, Kalen Michael, TaoXie, Jiacong Fang, imyhxy, Lorna, Zeng~Yifu(Zeng Yifu), Colin Wong, Abhiram V, Diego Montes, Zhiqiang Wang, Cristi Fati, Jebastin Nadar, Laughing, UnglvKitDe, Victor Sonck, tkianai, yxNONG, Piotr Skalski, Adam Hogan, Dhruv Nair, Max Strobel, and Mrinal Jain.
\newblock ultralytics/yolov5: v7.0 - {YOLOv5} {SOTA} {Realtime} {Instance} {Segmentation}, November 2022.
\newblock URL \url{https://zenodo.org/records/7347926}.

\bibitem[Dosovitskiy et~al.(2020)]{dosovitskiy2020image}
Alexey Dosovitskiy et~al.
\newblock An image is worth 16x16 words: Transformers for image recognition at scale.
\newblock \emph{arXiv preprint arXiv:2010.11929}, 2020.

\bibitem[Carreira and Zisserman(2017)]{carreira2017quo}
J.~Carreira and A.~Zisserman.
\newblock Quo vadis, action recognition? a new model and the kinetics dataset.
\newblock In \emph{2017 IEEE Conference on Computer Vision and Pattern Recognition (CVPR)}, 2017.

\bibitem[Carreira et~al.(2018)Carreira, Noland, Banki-Horvath, Hillier, and Zisserman]{carreira2018short}
J.~Carreira, E.~Noland, A.~Banki-Horvath, C.~Hillier, and A.~Zisserman.
\newblock A short note about kinetics-600.
\newblock \emph{CoRR}, 2018.

\bibitem[Bertasius et~al.(2021)Bertasius, Wang, and Torresani]{bertasius2021space}
Gedas Bertasius, Heng Wang, and Lorenzo Torresani.
\newblock Is space-time attention all you need for video understanding?
\newblock In \emph{ICML}, volume~2, 2021.

\end{thebibliography}

\end{document}